\documentclass{article}

\usepackage[nonatbib,final]{arxiv_nips_2016}

\usepackage[utf8]{inputenc} 
\usepackage[T1]{fontenc}    
\usepackage[colorlinks,citecolor=blue,linkcolor=blue]{hyperref}       
\usepackage{url}            
\usepackage{booktabs}       
\usepackage{amsfonts}       
\usepackage{nicefrac}       
\usepackage{microtype}      
\usepackage{verbatim}
\usepackage[svgnames]{xcolor}
\usepackage{amsmath,amssymb}
\usepackage{algorithm}
\usepackage{algorithmic}
\usepackage{graphicx}
\usepackage{subcaption}
\usepackage{tikz}
\usepackage{booktabs,chemformula}

\usetikzlibrary{shapes,arrows,arrows.meta}
\tikzset{%
  >={Latex[width=2mm,length=2mm]},
            base/.style = {rectangle, rounded corners, draw=black,
                           minimum width=1.5cm, minimum height=1cm,
                           text centered, font=\sffamily},
  activityStarts/.style = {base, fill=blue!30},
    InExtractPortion/.style = {base, fill=blue!30},
       InfoAggregator/.style = {base, fill=red!30},
    ColGen/.style = {base, fill=green!10},
    Storage/.style = {base, fill=brown!30},
    ILPSolve/.style = {base, fill=red!30},
    Output/.style = {base, fill=gray!30},
       startstop/.style = {base, fill=red!30},
    activityRuns/.style = {base, fill=green!30},
         process/.style = {base, minimum width=2.5cm, fill=orange!15,
                           font=\ttfamily},
}

\definecolor{Gray}{gray}{0.9}

\pgfdeclarelayer{edgelayer}
\pgfdeclarelayer{nodelayer}
\pgfsetlayers{edgelayer,nodelayer,main}

\tikzstyle{vertex_default}=[circle,fill=White,draw=Black,text width=0.1cm]
\tikzstyle{tree_edge}=[thick,draw=Green]
\tikzstyle{add_edge}=[thick,draw=Red]
\tikzstyle{local_edge}=[thick,draw=Cyan]

\newcommand\ie{\emph{i.e.}}
\newcommand\eg{\emph{e.g.}}

\title{Exploiting skeletal structure in computer vision annotation with Benders decomposition}

\title{Exploiting skeletal structure in computer vision annotation with Benders decomposition}
\author{
Shaofei Wang\\
Beijing A\&E Technologies \\
Beijing China \\
\texttt{sfwang0928@gmail.com} 
\And
Konrad Kording\\
University of Pennsylvania \\
Philadelphia PA  \\
\texttt{koerding@gmail.com} 
\And
Julian Yarkony\\
Experian Data Lab \\
San Diego CA \\
\texttt{julian.e.yarkony@gmail.com} 
}

\begin{document}
%
\maketitle
\begin{abstract}
Many annotation problems in computer vision can be phrased as integer linear programs (ILPs). The use of standard industrial solvers does not to exploit the underlying structure of such problems \eg, the skeleton in pose estimation. The leveraging of the underlying structure in conjunction with industrial solvers promises increases in both speed and accuracy. Such structure can be exploited using  Bender's decomposition, a technique from operations research, that solves complex ILPs or mixed integer linear programs by decomposing them into sub-problems that communicate via a master problem. The intuition is that conditioned on a small subset of the variables the solution to the remaining variables can be computed easily by taking advantage of properties of the ILP constraint matrix such as block structure. In this paper we apply Benders decomposition to a typical problem in computer vision where we have many sub-ILPs (\eg, partitioning of detections, body-parts) coupled to a master ILP (\eg, constructing skeletons). Dividing inference problems into a master problem and sub-problems motivates the development of a plethora of novel models, and inference approaches for the field of computer vision.  
\end{abstract}
\section{Introduction}
\subsection{Human Pose Estimation}
For a wide variety of applications in sports science, training, and rehabilitation, the quantification of human movement is crucial. While this is typically done in specialized labs, it is possible to estimate the three dimensional position of body parts (jointly referred to as a pose for a given person) of subjects from single camera video. 
Such problems are often modeled using an underlying tree structured model of the human body \cite{felzenszwalb2005pictorial} that models the following two components. The first softly enforces that part locations of a predicted person are supported by evidence in the image as described by local image features \cite{dalal2005histograms,vondrick2013hoggles}. The second component softly enforces that the part locations of a predicted person satisfy the angular and distance relationships consistent with a person \cite{deva3}. An example of such a relationship is that the head of a person tends to be above the neck. To deal with multiple people in an image, sliding window detectors\cite{viola2004robust,dalal2005histograms,felzenszwalb2010object,yang2012recognizing} are employed which check for people across multiple positions, scales, orientations etc.  These approaches, however, do not explicitly model the presence and joint likelihood of multiple people outside of tree structured models \cite{yang2012recognizing}, which is necessary for meaningful multi-person pose estimation(MPPE).  
%
\subsection{ ILP Formulation for MPPE}
MPPE has been modeled as a correspondence problem in \cite{deepcut2,deepcut1} where candidate detections (of body parts) are grouped to form each pose. This correspondence problem is formulated as an ILP that labels detections as either inactive or associated with a given part and then clusters the active detections into poses according the the multi-cut clustering criteria.  The objective and part detections for this ILP are produced by a deep neural network\cite{hinton2012deep,hinton} and a spring model for the distances between body parts. 
\subsection{Column Generation}
Column generation \cite{barnprice} is an efficient way of solving  linear programs (LPs) that have certain structural properties, namely a block structure exhibited by many LPs and ILPs including cutting stock \cite{gilmore1961linear}, routing vehicles\cite{desaulniers2006column}, facility location \cite{barahona1998plant}, crew scheduling\cite{vance1997airline}, and more recently in computer vision, MPPE\cite{wang2016efficient}, hierarchical image segmentation\cite{HPlanarCC}, and multi-object tracking \cite{yarkoNips2016}.  In such cases a new LP relaxation, that is often much tighter than a compact formulation \cite{geoffrion2010lagrangian,armacost2002composite}, is created that includes an enormous number of variables and a finite number of constraints.  

The corresponding optimization problem is solved by iteratively solving the dual LP with a limited set of the dual constraints followed by identifying violated dual constraints using combinatorial optimization (usually dynamic programming) which are then added to the set under consideration. This process terminates when no remaining dual constraints are violated at which point the corresponding primal LP solution is guaranteed to be optimal. The process of identifying violated dual constraints is commonly called ``pricing" in operations research.
\subsection{Two-Tier  Formulation of Multi-Person Pose Estimation}
\label{probForm}
An alternative model of the MPPE problem is the two-tier formulation (TTF)\cite{wang2016efficient}.  TTF can be understood as an approximation to \cite{deepcut2,deepcut1} such that exact or near exact inference can be achieved efficiently.  
TTF starts with detections of body parts from a deep convolutional neural network \cite{Deepseg,krizhevsky2012imagenet}
each of which is associated with  exactly one body part. Next TTF reasons about the assignment of parts to poses in two tiers:  \emph{local assignments},
which correspond to groupings of detections each of which is associated with a single part in a single  pose; and \emph{skeletons}, each of which models the outline of a pose and in which each part is associated with no more than one detection.

TTF then solves an ILP formulation over the skeletons and local assignments using costs derived identically to \cite{deepcut2,deepcut1}.  TTF is designed with inference via column generation in mind and leverages dynamic programming to construct skeletons and explicit enumeration for local assignments. %
\subsection{Benders Decomposition}
%
In order to exploit the structure of TTF we find further inspiration in operations research in the form of  Bender's decomposition\cite{benders1962partitioning}.  
In Benders decomposition, the variables of the original problem are divided into at least two subsets, one of which is identified with a LP (or ILP) called the master problem with the remainder identified with  LPs called sub-problems (also called Benders sub-problems).  Conditioned on the solution to the master problem each of the Benders problems is solved independently.  Optimization of the master problem relies on transforming the sub-problems to their dual forms and rewriting optimization over only variables associated with the master problem.  Solutions to the dual sub-problems form constraints in the master that are added on demand over the course of optimization.

Benders decomposition is especially powerful in domains in which both of the following two conditions hold.
\begin{itemize}
\item The LP relaxation of the ILP is inclined to produce fractional values for the variables of the  master problem.
\item Conditioned on an integer solution to the master problem, the variables of the sub-problems tend not to be fractional or are allowed to be fractional or non-negative.   
\end{itemize}
Such problems are common in the routing literature in which the master variables correspond to pipes, depots, or routes of vehicles while the sub-problem variables correspond to flows of  commodities  or passengers \cite{costa2005survey,cordeau2001benders,geoffrion1974multicommodity}.  An example is the problem of selecting a set of depots to open and jointly allocating flow of commodities from depots to clients.  In this case depots are associated with binary variables in the master problem and flows of commodities from depots is considered in the sub-problem.  The cost function trades off the fixed cost of opening depots and the variable cost of transporting commodities from depots to clients.

  We now consider a modification of the example above that is particularly amenable to Benders decomposition.  Consider that multiple commodities are modeled and that conditioned on the depots opened that routing of the commodities can be done independently.  In this case there is one Benders sub-problem for each commodity which is a highly desirable feature that accelerates optimization.  	

In our Benders decomposition approach we employ a master problem that determines the skeletons and one Benders sub-problem for each body part that associated the local assignments for that body part to the skeletons. 
\subsection{Our Approach}
In this paper we introduce Benders decomposition  to computer vision and apply it to optimization over the TTF formulation of MPPE. First 
we review the TTF formulation of \cite{wang2016efficient}.  Next 
we propose a Benders decomposition formulation for TTF. 
Then 
we contrast our approach to the approach of \cite{wang2016efficient}. 
Next we demonstrate efficiency of inference on problems from the MPPE benchmark \cite{andriluka14cvpr}.  Finally we conclude and discuss extensions.  
\section{The TTF Model}
\label{MathForm}
\paragraph{Body parts and the corresponding detections.}
Let the sets of detections, parts be denoted $\mathcal{D},\mathcal{R}$, and indexed by $d,r$ respectively.  We use $\mathcal{R}'$ to denote the subset of $\mathcal{R}$ corresponding to major parts of which at least one must be included in any given pose.
We describe the correspondence between detections and parts using a vector $R \in \{\mathcal{R}\}^{|\mathcal{D}|}$ which we index by $d$ where $R_{d}$ is the part associated with detection $d$.  
\paragraph{Skeletons.}
%
 We define the set of possible skeletons
over $\mathcal{D}$ as $\mathcal{G}$. Each element of $\mathcal{G}$ contains at least one detection corresponding to a major part and no more than one detection corresponding to any given part. Mappings of detections to skeletons are expressed by a matrix $G \in \{ 0,1\}^{|\mathcal{D}|\times |\mathcal{G}|}$, 
$G_{dg}=1\; \iff $ detection $d$ is included in skeleton $g$. 
We maintain an active set of skeletons, $\hat{\mathcal{G}}$, since $\mathcal{G}$ is impractically large.
\paragraph{Local Assignments.}

The set of possible local assignments
over the detections $\mathcal{D}$ is denoted $\mathcal{L}$, which we index by $l$.  
We describe $\mathcal{L}$ using
the matrices $L, M \in \{0,1\}^{|\mathcal{D}| \times |\mathcal{L}|}$,
where $L_{dl}=1$ $\iff$ detection $d$ is associated with $l$ as a local 
detection, and $M_{dl}=1$ $\iff$ detection $d$ is associated with
$l$ as a global detection. 
We maintain an active set of local assignments, $\hat{\mathcal{L}}$, since $\mathcal{L}$ is impractically large.
\paragraph{Validity Constraints.}
\label{validDef}
We use indicator vectors to describe a selection of skeletons and local assignments.
We use  $\gamma \in \{0,1\}^{|\mathcal{G}|}$ where 
$\gamma_g=1$ indicates that skeleton $g \in \mathcal{G}$ is selected, and 
$\gamma_g=0$ otherwise.  Similarly, we let $\psi\in \{0,1\}^{|\mathcal{L}|}$
where $\psi_l=1$ indicates that local
assignment $l\in \mathcal{L}$ is selected, with $\psi_l=0$ otherwise.
We describe the validity of pair ($\gamma,\psi$) using the following
linear inequalities:
\begin{itemize}
\item 
$-G\gamma+M\psi\leq 0$:  The global detection of a selected local assignment is included in some selected skeleton.
\item 
$L\psi+M\psi \leq 1$:  Detections can not be 
shared between selected local assignments.
\item 
$G\gamma+L\psi \leq 1$: A detection can only be 
global, local, or neither and can not be shared between selected skeletons.
\end{itemize}
\paragraph{Cost Function.}
\label{costPose}
We associate $\gamma,\psi$ with a cost function which is described using  $\theta \in \mathbb{R}^{|\mathcal{D}|}$,$\phi \in \mathbb{R}^{|\mathcal{D}|\times |\mathcal{D}|}$,$\omega \in \mathbb{R}$.
Here $\theta_d,\phi_{d_1d_2}$ are respectively the costs of assigning detection $d$ to a pose, and $d_1$ and $d_2$ to a common skeleton or local assignment. We use $\omega$ to denote the cost of instancing a pose which regularizes the number of people predicted.  
%
TTF associates costs to skeletons and local assignments with $\Gamma \in \mathbb{R}^{|\mathcal{G}|}$ and  $\Psi \in \mathbb{R}^{|\mathcal{L}|}$ where  $\Gamma_g,\Psi_l$  are the costs of skeleton $g$ and local assignment $l$ respectively, which are defined below. 
\begin{align}
\Gamma_g&=\omega+\sum_{d\in \mathcal{D}}\theta_dG_{dg}+\sum_{\substack{d_1\in \mathcal{D}\\d_2\in \mathcal{D}}}\phi_{d_1d_2}G_{d_1g}G_{d_2g}\\
\nonumber \Psi_{l}&=\sum_{d\in \mathcal{D}}\theta_d L_{dl} 
 + \sum_{\substack{d_1\in \mathcal{D}\\d_2\in \mathcal{D}}}\phi_{d_1d_2}(L_{d_1 l}+M_{d_1 l})(M_{d_2 l}+L_{d_2 l})
\end{align}
%
\paragraph{Integer Linear Program.}
\label{optDef}
TTF formulates MPPE as the following ILP which selects the lowest total cost valid set of skeletons, local assignments.
\begin{align}
\label{primdualilp}
\min_{\substack{\gamma \in \{0,1\}^{|\mathcal{G}|} \\ \psi \in \{0,1\}^{|\mathcal{L}|}}} & \Gamma^\top\gamma+\Psi^\top\psi  \\
\nonumber \text{s.t.  } \quad & G\gamma+L\psi \leq 1 \\\quad 
                        &\nonumber L\psi+M\psi\leq 1 \\
                        &\nonumber -G\gamma+M\psi\leq 0
\end{align}
\section{Benders Solution}
\label{BendersLP}
In this section we develop our Benders decomposition based solution for  optimization in the TTF model.  
 We begin by adding the redundant constraint to Eq \ref{primdualilp} that no global detection is associated with more than one selected skeleton ($G\gamma \leq 1$) then relaxing integrality.  
\begin{align}
\label{withext}
\mbox{Eq } \ref{primdualilp}\geq \min_{\substack{\gamma \geq 0\\ \psi \geq 0  \\ G\gamma \leq 1\\ G\gamma+L\psi \leq 1 \\ M\psi+L\psi \leq 1\\ M\psi-G\gamma \leq 0}}\Gamma^{\top}\gamma+\Psi^{\top}\psi
\end{align}
We use the $G^r,L^r,M^r$ to refer to the matrices corresponding to the subset of the rows of  $G,L,M$ over detections of part $r$.   Similarly $\psi^r,\Psi^r,\mathcal{L}^r$  concern all local assignments over part $r$.  We now rewrite optimization using  this separation.
\begin{align}
\label{withR}
\mbox{Eq } \ref{withext}= \min_{\substack{\gamma \geq 0\\G\gamma \leq 1}}\Gamma^{\top}\gamma+
\sum_{r \in \mathcal{R}} \min_{\substack{ \psi^r \geq 0 \\  G^r\gamma+L^r\psi^r \leq 1\\ M^r\psi^r+L^r\psi^r \leq 1 \\ M^r\psi^r-G^r\gamma \leq 0}}\Psi^{\top r}\psi^r
\end{align}
We refer to the optimizations over $\psi^r$ terms as sub-problems and  introduce their dual form using dual variables $\lambda^{1r},\lambda^{2r} ,\lambda^{3r} $ each of which lie in $\mathbb{R}_{0+}^{\mathcal{D}}$ for each $r\in \mathcal{R}$. 
\begin{align}
\label{sep2a}
\mbox{Eq } \ref{withR}= \min_{\substack{\gamma \geq 0 \\G\gamma \leq 1}}&\Gamma^{\top}\gamma+
 \sum_{r \in \mathcal{R}} \max_{\substack{ \lambda^{1r}\geq 0 \\ \lambda^{2r}\geq 0\\ \lambda^{3r}\geq 0 }} -1^{\top}\lambda^{1r}-1^{\top}\lambda^{2r}+\gamma^{\top}G^{r\top}\lambda^{1r}-\gamma^{\top}G^{r\top}\lambda^{3r}\\
 \nonumber \text{s.t. } \quad & \Psi^r+L^{r\top}\lambda^{1r}+M^{r\top}\lambda^{2r}+L^{r\top}\lambda^{2r}+M^{r\top}\lambda^{3r} \geq 0
\end{align}%
Notice that feasibility of the dual problem is not affected by the value of $\gamma$.  We now consider the set of all dual feasible solutions for sub-problem $r$ as $\Lambda^r$ and the union of such sets as $\Lambda$ and index such sets with $i$.  We use $\lambda^{i1r},\lambda^{i2r},\lambda^{i3r}$ to describe a member $i$ of $\Lambda^{r}$. We refer to elements in $\Lambda^r$ for any $r$ as Benders rows or rows for shorthand.  We now rewrite optimization using $-\ell_r$ to denote the objective of the sub-problem $r$.   Notice that the optimal sub-problem objective is always non-positive regardless of the skeletons chosen (thus $\ell_r$ must be non-negative).  
\begin{align}
\label{primalBender}
\mbox{Eq } \ref{sep2a}= \min_{\substack{\gamma \geq 0\\G\gamma \leq 1 \\ \ell_r \geq 0\quad r \in \mathcal{R}}}&\Gamma^{\top}\gamma+
\sum_{r \in \mathcal{R}} -\ell_r \\
\nonumber \text{s.t. } \quad &-\ell_r \geq -1^{\top}\lambda^{i1r}-1^{\top}\lambda^{i2r} +\gamma^{\top}G^{r\top}\lambda^{i1r}-\gamma^{\top}G^{r\top}\lambda^{i3r} \quad \forall r \in \mathcal{R},i \in \Lambda^r
\end{align} 
We write the dual form of Eq \ref{primalBender} below using Lagrange multipliers $\mu$.  Here $\mu$ has one index for each $d$ (denoted $\mu_d^0$) and one index for each (i,r) pair (denoted $\mu_{ir}$).
\begin{align}
\label{dualBender}
\mbox{Eq }\ref{primalBender}=\max_{\substack{\mu^0_d \geq 0\\ \mu_{ir} \geq 0  \\ \sum_{i \in \Lambda^r}\mu_{ir}\geq 1 }}&\sum_{d \in \mathcal{D}}- \mu_d^0+\sum_{\substack{r \in \mathcal{R}\\ i \in \Lambda^r}}\mu_{ir}(-1^{\top}\lambda^{i1r}-1^{\top}\lambda^{i2r})\\
\nonumber \\
\nonumber \text{s.t. } \quad &\Gamma_g+\sum_{d \in \mathcal{D}} \mu^0_dG_{dg} + 
\nonumber\sum_{\substack{r \in \mathcal{R}\\ i \in \Lambda^r}}\mu_{ir}\sum_{d \in \mathcal{D}}(\lambda^{i1r}_d-\lambda^{i3r}_d)G^{r}_{dg}\geq 0 \quad \forall g \in \mathcal{G}
\end{align}
%
\subsection{Solving Benders Formulation}
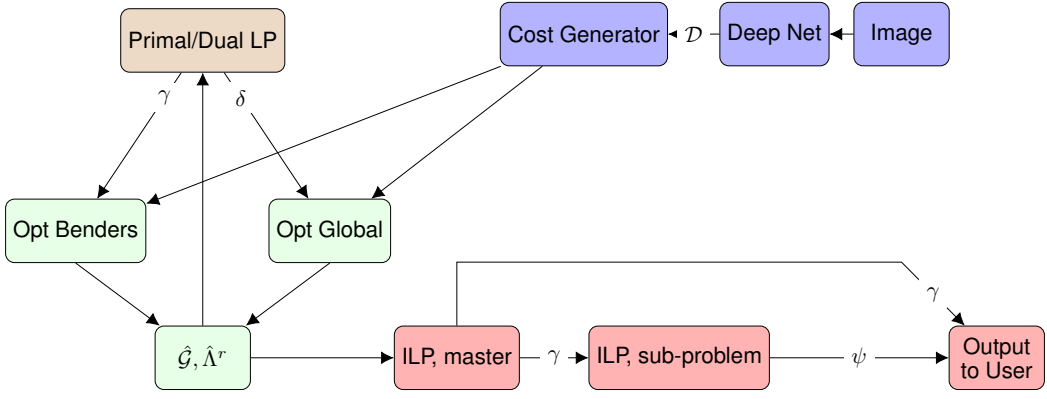
\begin{figure}[t]
\centering
\begin{subfigure}{1\textwidth}
\resizebox{\textwidth}{!}{
\begin{tikzpicture}[node distance=2cm,
    every node/.style={fill=white, font=\sffamily}, align=center]
%
 \node (rawImage)         [InExtractPortion]              {Image}; \\
 \node (DeepNet)          [InExtractPortion,left of =rawImage,xshift=-.001cm]              {Deep Net};\\
 \node (CRFGen)           [InExtractPortion,left of =DeepNet,xshift=-1cm]              {Cost Generator};
 \node (CostAgg)          [Storage,left of =CRFGen,yshift=-.1cm,xshift=-4cm]              {Primal/Dual LP};
 \node (Col1)             [ColGen,below of =CostAgg,yshift=-1cm,xshift=-2cm]              {Opt Benders};
 \node (Col2)             [ColGen,below of =CostAgg,yshift=-1cm,,xshift=2cm]              {Opt Global};
 \node (hatP)             [ColGen,below of =Col2,xshift=-2cm]              {$\hat{\mathcal{G}},\hat{\Lambda}^r$};
 \node (ILPSolve)         [ILPSolve,right of =hatP,xshift=2cm]              {ILP, master};
 \node (ILPSolve2)         [ILPSolve,right of =ILPSolve,xshift=1.5cm]              {ILP, sub-problem};
 \node (OutputModel)      [ILPSolve,right of =ILPSolve2,xshift=3cm]              {Output\\ to User};
 \draw[->]                (rawImage) --  (DeepNet);
 \draw[->]                (DeepNet) -- node {$\mathcal{D}$}(CRFGen);
 \draw[->]                (CRFGen) --  (Col1);
 \draw[->]                (CRFGen) --  (Col2);
 %
 
   \draw[->]              (CostAgg) -- node[pos=0.2] {$\gamma$}(Col1);
 \draw[->]                (CostAgg) -- node[pos=0.2] {$\delta$}(Col2);

 \draw[->]                (hatP) -- (ILPSolve);
   \draw[->]              (Col1.south) -- (hatP);
 \draw[->]                (Col2.south) -- (hatP);

 \draw[->]                (hatP) -- (CostAgg);

 \draw[->]                (ILPSolve.north) -- ++(0,1)-- ++(7,0) -- node {$\gamma$}(OutputModel);
 \draw[->]                (ILPSolve) -- node {$\gamma$}(ILPSolve2);
 \draw[->]                (ILPSolve2) -- node {$\psi$}(OutputModel);
\end{tikzpicture}
}
\end{subfigure}
\caption{Diagram of our Benders system: blue blocks represent steps for generating unary
  and pairwise costs. Green blocks represent steps for generating columns/rows.  \textbf{Opt
  Benders} and \textbf{Opt Global} correspond to the pricing problems.  The brown block
  represents a primal/dual LP solver while red blocks show steps for producing the final
  integer solutions at termination.}
\label{bendersFigure}
\end{figure}

Since we can not enumerate all possible Benders rows or skeletons we iteratively construct small subsets sufficient for optimality.  We use $\hat{\Lambda}^{r}$,$\hat{\mathcal{G}}$ to denote the nascent subsets of $\Lambda^{r}$,$\mathcal{G}$ and write the corresponding optimization below.   
\begin{align}
\label{limited}
\min_{\substack{\gamma \geq 0 \\G\gamma \leq 1 \\ \gamma_g=0 \forall g \notin \hat{\mathcal{G}} \\ \ell_r \geq 0}}&\Gamma^{\top}\gamma+
\sum_{r \in \mathcal{R}} -\ell_r \\
\nonumber \text{s.t. } \quad & -\ell_r \geq -1^{\top}(\lambda^{i1r}+\lambda^{i2r})+
 \gamma^{\top}G^{r\top}\lambda^{i1r}-\gamma^{\top}G^{r\top}\lambda^{i3r} \quad \forall r\in \mathcal{R},i \in \hat{\Lambda}^r \\
 =\nonumber  \max_{\substack{\mu^0_d \geq 0\\ \mu_{ir} \geq 0\\ \sum_{i \in \Lambda^r}\mu_{ir}\geq 1 \\ \mu_{ir}=0 \forall r\in \mathcal{R},i \notin \hat{\Lambda}^r}}&\sum_{d \in \mathcal{D}} \mu^0_d+\sum_{\substack{r \in \mathcal{R}\\ i \in \Lambda^r}}\mu_{ir}(-1^{\top}\lambda^{i1r}-1^{\top}\lambda^{i2r})\\
\nonumber \text{s.t. } \quad & \Gamma_g+\sum_{d \in \mathcal{D}} \mu^0_dG_{dg} +
\nonumber \sum_{\substack{r \in \mathcal{R}\\ i \in \Lambda^r}}\mu_{ir}\sum_{d \in \mathcal{D}}(\lambda^{i1r}_d-\lambda^{i3r}_d)G^{r}_{dg}\geq 0 \quad \forall g \in \hat{\mathcal{G}}
\end{align} 
We now formally study the solution procedure for the primal and dual form in Eq \ref{limited} which are solved jointly.  We diagram our procedure in Fig \ref{bendersFigure}. For each $r\in \mathcal{R}$ we initialize $\hat{\Lambda}^r$ with the solution corresponding to $\gamma_g=0$ for all $g$ and initialize $\hat{\mathcal{G}}$ with the empty set.  We then proceed by solving the LP in Eq \ref{limited} producing both the primal solution and the dual solution.  We use the dual solution to find violated dual constraints corresponding to $\mathcal{G}$ and the primal solution to find violated constraints in $\Lambda^r$ for all $r \in \mathcal{R}$.  We repeat these steps until convergence.  We study the generation of members of $\hat{\mathcal{G}}$ and $\hat{\Lambda}^r$  in the succeeding subsections.  
\begin{algorithm}[tb]
\caption{Benders Decomposition Dual Optimization }
\begin{algorithmic}[1]
\STATE $\hat{\mathcal{G}} \leftarrow \emptyset$
\STATE $\hat{\mathcal{L}}^r \leftarrow \emptyset, \forall r \in \mathcal{R}$
\STATE $\hat{\Lambda^r} \leftarrow \{  $ $i \leftarrow \max_{\substack{ \lambda^{1r}\geq 0 \\ \lambda^{2r}\geq 0\\ \lambda^{3r}\geq 0}}-1^{\top}\lambda^{1r}-1^{\top}\lambda^{2r}$ over dual feasible space $\}$
\REPEAT 
\STATE did\_change$\leftarrow 0$
\STATE $[\lambda,\gamma] \leftarrow$ Maximize primal/dual in Eq.~\eqref{dualFormPose} over column sets $\hat{\mathcal{G}},\hat{\Lambda}^r$
\FOR {$d_{*} \in \mathcal{D}$  s.t. $ R_{d_*}$  $ \in \mathcal{R}'$}
\STATE $g_* \leftarrow \mbox{arg} \min_{\substack{g \in \mathcal{G} \\ G_{d_*g}=1}}\Gamma_g+\sum_{d \in \mathcal{D}} G_{dg}\delta_d$
\IF{$\Gamma_{g_*}+\sum_{d \in \mathcal{D}} G_{dg_*}\delta_d<0$}
\STATE $\hat{\mathcal{G}} \leftarrow [\hat{\mathcal{G}} \cup g_*]$
\STATE did\_change$\leftarrow 1$
\ENDIF
\ENDFOR
\FOR {$r \in \mathcal{R}$}
\REPEAT
\STATE $\lambda^r_* \leftarrow \max_{\substack{ \lambda^{1r}\geq 0 \\ \lambda^{2r}\geq 0\\ \lambda^{3r}\geq 0}}-1^{\top}\lambda^{1r}-1^{\top}\lambda^{2r}+\gamma^{\top}G^{r\top}\lambda^{1r}-\gamma^{\top}G^{r\top}\lambda^{3r}$ over dual feasible space
\STATE $\dot{\mathcal{L}}^r \leftarrow \emptyset$
\FOR {$d_{*} \in \mathcal{D}$  s.t. $R_{d_*} = r$}
\STATE $l_* \leftarrow \mbox{arg} \min_{\substack{l\in \mathcal{L}^r \\ M^r_{d_*l}=1}}
(\lambda^{3r}_{d_*}+\lambda^{2r}_{d_*})M^r_{d_*l}+\sum_{d \in \mathcal{D}}(\lambda^{1r}_d+\lambda^{2r}_d)L^r_{dl}+\Psi^r_l$
\IF {$(\lambda^{3r}_{d_*}+\lambda^{2r}_{d_*})M^r_{d_*l_*}+\sum_{d \in \mathcal{D}}(\lambda^{1r}_d+\lambda^{2r}_d)L^r_{dl_*}+\Psi^r_{l_*}<0$}
\STATE $\dot{\mathcal{L}} \leftarrow [\dot{\mathcal{L}} \cup l_*]$
\STATE did\_change$\leftarrow 1$
\ENDIF
\ENDFOR
\STATE  $\hat{\mathcal{L}}^r \leftarrow [\hat{\mathcal{L}}^r,\dot{\mathcal{L}}^r]$
\UNTIL { $|\dot{\mathcal{L}}^r|=0 $}
\STATE   $i\leftarrow  \mbox{ index of } \Lambda^{r}\mbox{ corresponding to final } \lambda^r_*$
\STATE $\hat{\Lambda}^r \leftarrow [\hat{\Lambda}^r \cup i]$
\ENDFOR
 \UNTIL{ did\_change=0}
\end{algorithmic}
\label{dualBenders}
\end{algorithm}

\subsection{Constructing Nascent Set  $\hat{\mathcal{G}}$}
\label{colMake}
We consolidate the dual terms as follows using $\delta \in \mathbb{R}^{|\mathcal{D}|}$ defined as follows for index $d$.
\begin{align}
\delta_d=\mu^0_d+\sum_{i \in \Lambda^{R_d}}\mu_{ir}(\lambda^{i1r}_d-\lambda^{i3r}_d)
\end{align}
For each detection $d_*$ such that $R_{d_*} \in \mathcal{R}'$ (\ie, $d_*$ corresponds
to a major part), we compute the most violated constraint corresponding to a
skeleton that includes detection $d_*$ as follows. We use indicator vector
$x\in \{0,1\}^{|\mathcal{D}|}$, and define a new column $g$ to be included in $\hat{\mathcal{G}}$,
defined by $G_{dg} = x^*_d$ for all $d \in \mathcal{D}$, where $x^*$ is the solution to:
\begin{align}
  &\min_{\substack{g \in \mathcal{G} \\ G_{d_*g}=1}} \Gamma_g+\sum_{d \in \mathcal{D}} G_{dg}\delta_d\\
  \nonumber =& \omega + \min_{ \substack{x \in \{ 0,1\}^{|\mathcal{D}|} \\ x_{d_*}=1\\  \sum_{d \in \mathcal{D}}R_{dr}x_d \leq 1 \; \; \forall r \in \mathcal{R}}} \!\!\!\!\!\!\! \sum_{ d\in \mathcal{D}} (\theta_{d}+\delta_d)x_{d}
 +\sum_{d_1d_2 \in \mathcal{D}} \phi_{d_1d_2}x_{d_1}x_{d_2}
\end{align}
We employ a tree structured model augmented with connections from the neck (which is the only major part) to all other non-adjacent body parts.  Conditioned on the global detection associated with the neck, the conditional model is tree-structured and can be optimized using dynamic programming.
\subsection{Constructing Nascent Set $\hat{\Lambda}^r$}
\label{rowmake}
Generating the most violated Benders row conditioned on $\gamma$ for a given $r$ is a simple linear programming problem.   
\begin{align}
\label{rowGenLp}
\max_{\substack{ \lambda^{1r}\geq 0 \\ \lambda^{2r}\geq 0\\ \lambda^{3r}\geq 0}}&-1^{\top}\lambda^{1r}-1^{\top}\lambda^{2r}+\gamma^{\top}G^{r\top}\lambda^{1r}-\gamma^{\top}G^{r\top}\lambda^{3r}\\
\nonumber \text{s.t.} \quad & \Psi^r+L^{r\top}\lambda^{1r}+M^{r\top}\lambda^{2r}+L^{r\top}\lambda^{2r}+M^{r\top}\lambda^{3r} \geq 0
\end{align}
We apply this optimization for each $r$ and apply column generation to produce local assignments to solve the LP.  Notice that this decouples from the master problem any need to consider the operations in the sub-problems.   

The set of local assignments for any part $r$ is too large to consider in optimization but not too large to enumerate in advance.  Thus to solve Eq \ref{rowGenLp} we employ an explicit column generation approach where we iteratively solve the dual LP with a limited set of constraints and then identify violated constraints, which are added to the set under consideration.
\subsubsection{Crucial Point in  Optimization}
When solving optimization  in Eq \ref{rowGenLp} we add a tiny negative bias to objective corresponding to terms  $\lambda^{1r},\lambda^{3r}$.  This ensures that the corresponding terms do not increase beyond what is needed to produce an optimal dual solution.  

The additional small biases can be understood as ensuring that   $\lambda^{1r}_d-\lambda^{3r}_d$ becomes the marginal cost for adding a local assignment to the solution in which $d$ is the global detection.  This addition does not affect the optimality of the solution if the bias is small enough in magnitude. 
\subsection{Anytime Lower Bounds}
In the Appendix we consider the production of anytime lower bounds in the master problem and Benders sub-problems.  We write the corresponding bounds below and leave their derivations to the Appendix. 

We write below a lower bound corresponding to Benders sub-problem $r$ given any  non-negative $\lambda$.  
\begin{align}
\lambda^{1\top}(-1+G^r\gamma)-\lambda^{2\top}1-\lambda^{3\top}G^r\gamma+\sum_{\substack{d \in \mathcal{D}  \\ R_d=r}}\min[0,\min_{\substack{l\in \mathcal{L} \\ M_{dl}=1 }}\Psi_l]
\end{align}

We write below a lower  bound corresponding to the master problem given any set of Benders rows and Lagrange multipliers $\lambda$.
\begin{align}
 -\mu^{0\top}1 - \sum_{\substack{r \in \mathcal{R} \\ i \in \Lambda^r}}\mu_{ir}(1^{\top}(\lambda^{i1r}+\lambda^{i2r}))
+\sum_{\hat{d} \in \mathcal{R}'}\min_{\substack{g \in \mathcal{G} \\ G_{\hat{d}g}=1}}\gamma_g(\Gamma_g+\sum_{d \in \mathcal{D}}G_{dg}(\mu^0_d+\sum_{\substack{r \in \mathcal{R} \\ i \in \Lambda^r}}\mu_{ir}( \lambda^{i1r}_d-\lambda^{i3r}_d)))
\end{align}

\section{Comparison to Column Generation Formulation for TTF}
\label{comparison}
Our approach can be contrasted with the approach of \cite{wang2016efficient} which employs an optimization approach that we refer to as pure column generation (PCG) in contrast to our approach which we refer to as Benders Column generation (BCG).  PCG relaxing the integrality constraints on $\gamma,\psi$  producing the following primal/dual optimization over $\lambda^1,\lambda^2,\lambda^3 \in \mathbb{R}_{0+}^{|\mathcal{D}|}$:  
\begin{align}
\label{dualFormPose}
\min_{\substack{\gamma\geq 0 \\ \psi \geq 0 \\ G\gamma+L\psi \leq 1 \\ L\psi+M\psi\leq 1 \\-G\gamma+M\psi\leq 0}}\Gamma^\top\gamma+\Psi^\top \psi 
  =\max_{\substack{\lambda^1\geq 0\\ \lambda^2\geq 0 \\ \lambda^3\geq 0 \\  \Gamma+G^\top(\lambda^1-\lambda^3) \geq 0 \\ \Psi+L^\top\lambda^1+(M^\top+L^\top)\lambda^2+M^\top\lambda^3 \geq 0}} -1^\top\lambda^1-1^\top\lambda^2
\end{align}

Optimization in Eq \ref{dualFormPose} consists of iteratively solving the  dual form of Eq \ref{dualFormPose} using a limited set of the primal variables followed by identifying violated dual constraints.  This identification is done using dynamic programming and explicit column generation as in BCG.  

The BCG and PCG strategies attack the same LP relaxation and thus achieve the same objective.  BCG however is distinct in that it solves the optimizations associated with local assignments entirely separately from the master problem.  In contrast PCG reasons about both local assignments and skeletons along with all constraints in a common LP.  Both however use identical mechanisms to identify skeletons and local assignments that are added to consideration during optimization.   

\section{Experiments}
\label{exper}
We evaluate our approach against the baseline of \cite{wang2016efficient} using identical problem instances.  These problem instances are derived from that of \cite{deepcut1} and provided generously by its authors of \cite{wang2016efficient}. 

We have capped computation time at 300 seconds of dual optimization.  Afterwords we solve the ILP over the primal variables generated.  

\begin{figure}
  \centering
\includegraphics[width=9cm,trim={1cm 5cm 2cm 5cm},clip]{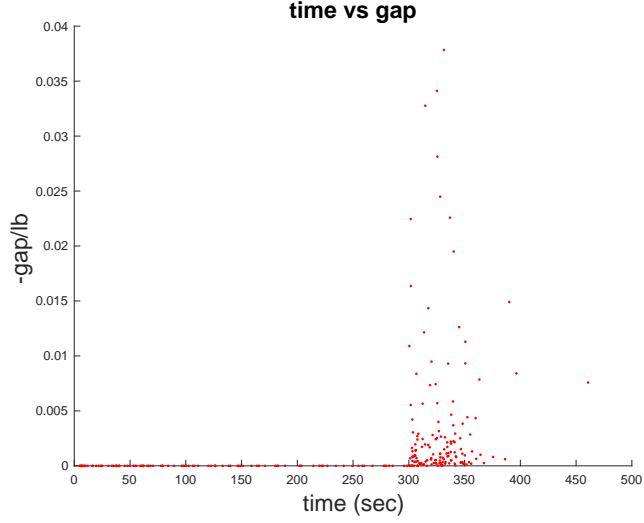}
\caption{Scatter plot of the normalized gap vs time}
\label{plotFig}
\end{figure}

\begin{figure}
  \centering
\includegraphics[width=9cm,trim={1cm 5cm 2cm 5cm},clip]{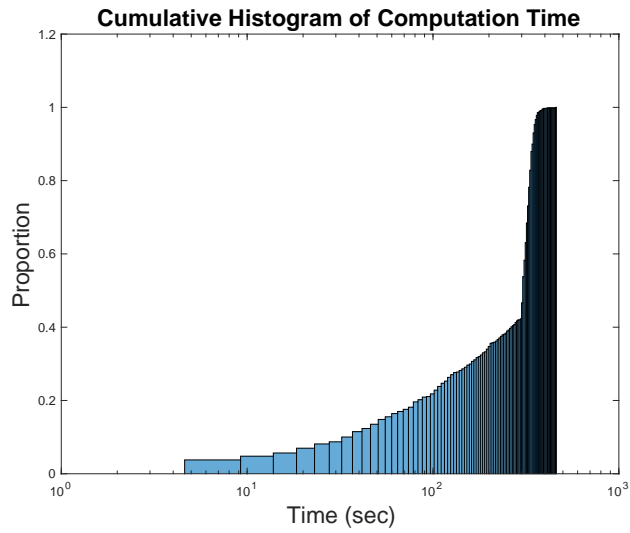}
\caption{Cumulative histogram displaying proportion of problems with consuming  less than a given amount of time.    }
\label{plotFigTime}
\end{figure}

\begin{figure}
  \centering
\includegraphics[width=9cm,trim={1cm 5cm 2cm 5cm},clip]{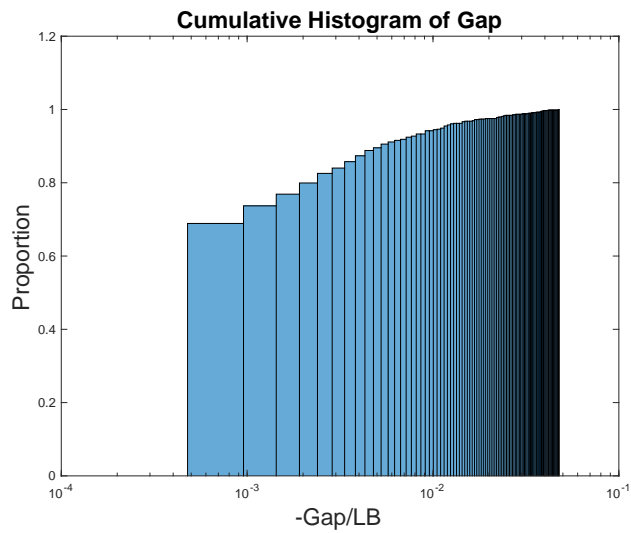}
\caption{Cumulative histogram of proportion of problems with gap less than a given value. }
\label{plotFigGap}
\end{figure}

We show a scatter plot of the gap between the upper and lower bound vs termination time in Fig \ref{plotFig}.  To normalize these gaps we divide by the minus one times maximum lower bound produced at termination for each problem instance.  We also plot a histogram of the computation time and gap in Fig \ref{plotFigTime} and Fig \ref{plotFigGap} respectively.  

We have results on  688 problem instances currently.  We observed some numerical issues in the LP solver that we do not see in PCG.  All experiments are conducted with MATLAB 2016 with its built in linprog LP solver using a 2014 MacBook Pro with 2.6 GHz Intel Core i5 chip.  

\section{Conclusions}
\label{conc}
To meaningfully use the structural information contained in the skeleton of humans, we have attacked optimization in the TTF  using Benders decomposition \cite{benders1962partitioning}.  We reformulate the TTF LP relaxation to separate optimization over of skeletons and local assignments. Using our Benders decomposition approach we produce one separate optimization for each part as opposed to one single  sub-problem as is common in the Benders decomposition literature \cite{magnanti1981accelerating,costa2005survey,cordeau2001benders}.  We produce tight bounds and fast inference.  

The introduction of Bender's decomposition opens a number of new opportunities for computer vision. The decomposition allows the separation of the integer and non-integer components in mixed integer programs, which promises to help circumvent difficulties induced by relaxing the integer variables. This could help for solving tracking problems in which splits occur (as in cell division in cell tracking). It also allows problems with heavy nesting as in planar ultrametrics \cite{HPlanarCC} in which the master problem of one layer is the subproblem of the next. 

Similar approaches based on the nested Benders decomposition could allow for dynamic programs to be solved in which the state space for individual variables is tractable but not the vector product. This could have use when inference in high tree width problems are considered . Such a case would be in pose estimation where hyper-nodes (nodes in the high tree width representation) correspond to parts and their components are detections.

Lastly, the use of Benders decomposition promises to enable parallelization of tracking and annotation by permitting the sub-problems to be solved independently while at the same time not relying on a Lagrangian relaxation with sub-gradient optimization \cite{komodakis2011mrf,huisman2005combining}. 
%
\bibliographystyle{ieee}

\bibliography{bib_inst_track}

\begin{thebibliography}{10}\itemsep=-1pt

\bibitem{andriluka14cvpr}
M.~Andriluka, L.~Pishchulin, P.~Gehler, and B.~Schiele.
\newblock 2d human pose estimation: New benchmark and state of the art
  analysis.
\newblock In {\em Proceedings of the IEEE Conference on Computer Vision and
  Pattern Recognition}, pages 3686--3693, 2014.

\bibitem{armacost2002composite}
A.~P. Armacost, C.~Barnhart, and K.~A. Ware.
\newblock Composite variable formulations for express shipment service network
  design.
\newblock {\em Transportation science}, 36(1):1--20, 2002.

\bibitem{barahona1998plant}
F.~Barahona and D.~Jensen.
\newblock Plant location with minimum inventory.
\newblock {\em Mathematical Programming}, 83(1):101--111, 1998.

\bibitem{barnprice}
C.~Barnhart, E.~L. Johnson, G.~L. Nemhauser, M.~W.~P. Savelsbergh, and P.~H.
  Vance.
\newblock Branch-and-price: Column generation for solving huge integer
  programs.
\newblock {\em Operations Research}, 46:316--329, 1996.

\bibitem{benders1962partitioning}
J.~F. Benders.
\newblock Partitioning procedures for solving mixed-variables programming
  problems.
\newblock {\em Numerische mathematik}, 4(1):238--252, 1962.

\bibitem{cordeau2001benders}
J.-F. Cordeau, G.~Stojkovi{\'c}, F.~Soumis, and J.~Desrosiers.
\newblock Benders decomposition for simultaneous aircraft routing and crew
  scheduling.
\newblock {\em Transportation science}, 35(4):375--388, 2001.

\bibitem{costa2005survey}
A.~M. Costa.
\newblock A survey on benders decomposition applied to fixed-charge network
  design problems.
\newblock {\em Computers \& operations research}, 32(6):1429--1450, 2005.

\bibitem{dalal2005histograms}
N.~Dalal and B.~Triggs.
\newblock Histograms of oriented gradients for human detection.
\newblock In {\em Computer Vision and Pattern Recognition, 2005. CVPR 2005.
  IEEE Computer Society Conference on}, volume~1, pages 886--893. IEEE, 2005.

\bibitem{desaulniers2006column}
G.~Desaulniers, J.~Desrosiers, and M.~M. Solomon.
\newblock {\em Column generation}, volume~5.
\newblock Springer Science \& Business Media, 2006.

\bibitem{Deepseg}
C.~Farabet, C.~Couprie, L.~Najman, and Y.~LeCun.
\newblock Learning hierarchical features for scene labeling.
\newblock {\em IEEE Transactions on Pattern Analysis and Machine Intelligence},
  35(8):1915--1929, 2013.

\bibitem{felzenszwalb2010object}
P.~F. Felzenszwalb, R.~B. Girshick, D.~McAllester, and D.~Ramanan.
\newblock Object detection with discriminatively trained part-based models.
\newblock {\em IEEE transactions on pattern analysis and machine intelligence},
  32(9):1627--1645, 2010.

\bibitem{felzenszwalb2005pictorial}
P.~F. Felzenszwalb and D.~P. Huttenlocher.
\newblock Pictorial structures for object recognition.
\newblock {\em International journal of computer vision}, 61(1):55--79, 2005.

\bibitem{geoffrion2010lagrangian}
A.~M. Geoffrion.
\newblock Lagrangian relaxation for integer programming.
\newblock {\em 50 Years of Integer Programming 1958-2008}, pages 243--281,
  2010.

\bibitem{geoffrion1974multicommodity}
A.~M. Geoffrion and G.~W. Graves.
\newblock Multicommodity distribution system design by benders decomposition.
\newblock {\em Management science}, 20(5):822--844, 1974.

\bibitem{gilmore1961linear}
P.~C. Gilmore and R.~E. Gomory.
\newblock A linear programming approach to the cutting-stock problem.
\newblock {\em Operations research}, 9(6):849--859, 1961.

\bibitem{hinton2012deep}
G.~Hinton, L.~Deng, D.~Yu, G.~E. Dahl, A.-r. Mohamed, N.~Jaitly, A.~Senior,
  V.~Vanhoucke, P.~Nguyen, T.~N. Sainath, et~al.
\newblock Deep neural networks for acoustic modeling in speech recognition: The
  shared views of four research groups.
\newblock {\em IEEE Signal Processing Magazine}, 29(6):82--97, 2012.

\bibitem{huisman2005combining}
D.~Huisman, R.~Jans, M.~Peeters, and A.~P. Wagelmans.
\newblock Combining column generation and lagrangian relaxation.
\newblock {\em Column generation}, pages 247--270, 2005.

\bibitem{deepcut1}
E.~Insafutdinov, L.~Pishchulin, B.~Andres, M.~Andriluka, and B.~Schiele.
\newblock Deepercut: A deeper, stronger, and faster multi-person pose
  estimation model.
\newblock In {\em ECCV}, 2016.
\newblock (Accepted).

\bibitem{komodakis2011mrf}
N.~Komodakis, N.~Paragios, and G.~Tziritas.
\newblock Mrf energy minimization and beyond via dual decomposition.
\newblock {\em IEEE transactions on pattern analysis and machine intelligence},
  33(3):531--552, 2011.

\bibitem{krizhevsky2012imagenet}
A.~Krizhevsky, I.~Sutskever, and G.~E. Hinton.
\newblock Imagenet classification with deep convolutional neural networks.
\newblock In {\em Advances in neural information processing systems}, pages
  1097--1105, 2012.

\bibitem{magnanti1981accelerating}
T.~L. Magnanti and R.~T. Wong.
\newblock Accelerating benders decomposition: Algorithmic enhancement and model
  selection criteria.
\newblock {\em Operations research}, 29(3):464--484, 1981.

\bibitem{deepcut2}
L.~Pishchulin, E.~Insafutdinov, S.~Tang, B.~Andres, M.~Andriluka, P.~Gehler,
  and B.~Schiele.
\newblock {DeepCut}: Joint subset partition and labeling for multi person pose
  estimation.
\newblock In {\em CVPR}, 2016.

\bibitem{hinton}
D.~E. Rumelhart, G.~E. Hinton, and R.~J. Williams.
\newblock Parallel distributed processing: Explorations in the microstructure
  of cognition, vol. 1.
\newblock chapter Learning Internal Representations by Error Propagation, pages
  318--362. MIT Press, Cambridge, MA, USA, 1986.

\bibitem{vance1997airline}
P.~H. Vance, C.~Barnhart, E.~L. Johnson, and G.~L. Nemhauser.
\newblock Airline crew scheduling: A new formulation and decomposition
  algorithm.
\newblock {\em Operations Research}, 45(2):188--200, 1997.

\bibitem{viola2004robust}
P.~Viola and M.~J. Jones.
\newblock Robust real-time face detection.
\newblock {\em International journal of computer vision}, 57(2):137--154, 2004.

\bibitem{vondrick2013hoggles}
C.~Vondrick, A.~Khosla, T.~Malisiewicz, and A.~Torralba.
\newblock Hoggles: Visualizing object detection features.
\newblock In {\em Proceedings of the IEEE International Conference on Computer
  Vision}, pages 1--8, 2013.

\bibitem{yarkoNips2016}
S.~Wang, S.~Wolf, C.~Fowlkes, and J.~Yarkony.
\newblock Tracking objects with higher order interactions using delayed column
  generation.
\newblock {\em Artificial Intelligence and Statistics}, 2017.

\bibitem{wang2016efficient}
S.~Wang, C.~Zhang, M.~A. Gonzalez-Ballester, and J.~Yarkony.
\newblock Efficient pose and cell segmentation using column generation.
\newblock {\em arXiv preprint arXiv:1612.00437}, 2016.

\bibitem{yang2012recognizing}
Y.~Yang, S.~Baker, A.~Kannan, and D.~Ramanan.
\newblock Recognizing proxemics in personal photos.
\newblock In {\em Computer Vision and Pattern Recognition (CVPR), 2012 IEEE
  Conference on}, pages 3522--3529. IEEE, 2012.

\bibitem{deva3}
Y.~Yang and D.~Ramanan.
\newblock Articulated pose estimation with flexible mixtures-of-parts.
\newblock In {\em Computer Vision and Pattern Recognition (CVPR), 2011 IEEE
  Conference on}, pages 1385--1392. IEEE, 2011.

\bibitem{HPlanarCC}
J.~Yarkony and C.~Fowlkes.
\newblock Planar ultrametrics for image segmentation.
\newblock In {\em Neural Information Processing Systems}, 2015.

\end{thebibliography}
\pagebreak
\appendix 
\section{Anytime Lower Bounds on the Master Problem}
\label{lowerBSect}
Given any set of the Benders rows we produce any anytime lower bound the optimum of the LP relaxation of the master problem as follows.  
\begin{align}
\label{boundStep1}
\mbox{Eq } \ref{sep2a} \geq 
 \min_{\substack{\gamma \geq 0\\G\gamma \leq 1 \\ \ell_r \geq 0\quad r \in \mathcal{R}}}\Gamma^{\top}\gamma+
\sum_{r \in \mathcal{R}} -\ell_r \\
\nonumber -\ell_r \geq -1^{\top}\lambda^{i1r}-1^{\top}\lambda^{i2r}+\gamma^{\top}G^{r\top}\lambda^{i1r}-\gamma^{\top}G^{r\top}\lambda^{i3r} \quad \forall r \in \mathcal{R},i \in \hat{\Lambda}^r
\end{align}
We now convert the constraints into dual variables and but retain $G\gamma  \leq 1$ in the master problem.  
\begin{align}
\mbox{Eq } \ref{boundStep1}=\min_{\substack{\gamma \geq 0\\ G\gamma \leq 1 \\ \ell_r \geq 0\quad r \in \mathcal{R}}} \max_{\mu \geq 0}\Gamma^{\top}\gamma+
\sum_{r \in \mathcal{R}} -\ell_r \\
\nonumber +\mu^{0\top}(G\gamma-1)
\nonumber +\sum_{r\in \mathcal{R},i \in \hat{\Lambda}^r}\mu_{ir}(\ell_r -1^{\top}\lambda^{i1r}-1^{\top}\lambda^{i2r}+\gamma^{\top}G^{r\top}\lambda^{i1r}-\gamma^{\top}G^{r\top}\lambda^{i3r} )
\end{align}
We now relax the optimality of  $\mu$ and consider any dual feasible  $\mu$  which implies $\mu$ is non negative and that for each $r\in \mathcal{R}$ that  $ \sum_{i \in \Lambda^r}\mu_{ir}\geq 1$.  
\begin{align}
\label{boundStep2}
\mbox{Eq } \ref{boundStep1} \geq 
\min_{\substack{\gamma \geq 0 \\ G\gamma \leq 1\\ \ell_r \geq 0\quad r \in \mathcal{R}}} \Gamma^{\top}\gamma+
\sum_{r \in \mathcal{R}} -\ell_r \\
\nonumber +\mu^{0\top}(G\gamma-1)+
\nonumber \sum_{r\in \mathcal{R},i \in \hat{\Lambda}^r}\mu_{ir}(\ell_r -1^{\top}\lambda^{i1r}-1^{\top}\lambda^{i2r}+\gamma^{\top}G^{r\top}\lambda^{i1r}-\gamma^{\top}G^{r\top}\lambda^{i3r} )\\
\nonumber=\min_{\substack{\gamma \geq 0\\ G\gamma \leq 1\\ \ell_r \geq 0\quad r \in \mathcal{R}}} \Gamma^{\top}\gamma+\sum_{r \in \mathcal{R}} -\ell_r  -\mu^{0\top}1 + \sum_{r,i \in \hat{\Lambda}^r}\mu_{ir}(\ell^{r}-1^{\top}(\lambda^{i1r}+\lambda^{i2r}))\\
\nonumber +\sum_{g \in \mathcal{G}} \gamma_g(\Gamma_g+\sum_{d \in \mathcal{D}}G_{dg}(\mu^0_d+\sum_{r\in \mathcal{R},i \in \hat{\Lambda}^r}\mu_{ir}( \lambda^{i1r}_d-\lambda^{i3r}_d)))
\end{align}
Since $ \sum_{i \in \Lambda^r}\mu_{ir}\geq 1$ we know that $\ell_r$ is zero valued. 
\begin{align} 
\mbox{Eq }\ref{boundStep2}=\min_{\substack{\gamma \geq 0\\ G\gamma \leq 1}} \Gamma^{\top}\gamma -\mu^{0\top}1 - \sum_{r\in \mathcal{R},i \in \hat{\Lambda}^r}\mu_{ir}1^{\top}(\lambda^{i1r}+\lambda^{i2r})\\
\nonumber +\sum_{g \in \mathcal{G}} \gamma_g(\Gamma_g+\sum_{d \in \mathcal{D}}G_{dg}(\mu^0_d+\sum_{r\in \mathcal{R},i \in \hat{\Lambda}^r}\mu_{ir}( \lambda^{i1r}_d-\lambda^{i3r}_d)))
\end{align}
We now relax optimization to permit the selection  of one skeleton associated with each detection associated with a global part.  Note that in the case of multiple global parts a given detection may be include more than once.  
\begin{align}
\mbox{Eq }\ref{boundStep2}\geq 
 -\mu^{0\top}1 - \sum_{\substack{r \in \mathcal{R} \\ i \in \Lambda^r}}\mu_{ir}(1^{\top}(\lambda^{i1r}+\lambda^{i2r}))\\
\nonumber +\sum_{\hat{d} \in \mathcal{R}'}\min_{\substack{g \in \mathcal{G} \\ G_{\hat{d}g}=1}}\gamma_g(\Gamma_g+\sum_{d \in \mathcal{D}}G_{dg}(\mu^0_d+\sum_{\substack{r \in \mathcal{R} \\ i \in \Lambda^r}} \mu_{ir}( \lambda^{i1r}_d-\lambda^{i3r}_d)))
\end{align}
Note that the minimization over $g$ is simply the minimization done for generating columns and is known to be a dynamic program.  

\section{Anytime Lower Bounds on a Benders Sub-Problem}
We now consider the production of anytime lower bounds on the solution the the LP corresponding to a given Benders sub-problem.  The production of these allows for an anytime dual feasible solution to be generated prior to convergence of the LP of the dual sub-problem and hence anytime generation of Benders rows.  
Recall the dual optimization corresponding to part $r$ given master problem solution $\gamma$.  

\begin{align}
 \min_{\substack{ \psi^r \geq 0 \\  G^r\gamma+L^r\psi^r \leq 1\\ M^r\psi^r+L^r\psi^r \leq 1 \\ M^r\psi^r-G^r\gamma \leq 0}}\Psi^{\top r}\psi^r
\end{align}
We begin by augment our primal ILP with the constraint that $M^r\psi^r\leq 1$  and then add the Lagrange multipliers $\lambda^1,\lambda^2,\lambda^3$ to optimization.  
\begin{align}
 \min_{\substack{ \psi^r \geq 0 \\  M^r\psi^r\leq 1 }}\max_{\lambda \geq 0}\Psi^{\top r}\psi^r+\lambda^{1\top}(-1+G^r\gamma+L^r\psi^r)+\lambda^{2\top}(-1+G^r\gamma+L^r\psi^r )+\lambda^{3\top}(M^r\psi^r-G^r\gamma)
\end{align}
We now relax optimality in $\lambda$ producing the following bound.
\begin{align}
 \min_{\substack{ \psi^r \geq 0\\ M^r\psi^r \leq 1}}\Psi^{\top r}\psi^r+\lambda^{1\top}(-1+G^r\gamma)-\lambda^{2\top}1-\lambda^{3\top}G^r\gamma+(\lambda^{1}+\lambda^{2})^{\top}L^r\psi^r+(\lambda^{2}+\lambda^{3})^{\top}M^r\psi^r\\
 \nonumber= \lambda^{1\top}(-1+G^r\gamma)-\lambda^{2\top}1-\lambda^{3\top}G^r\gamma+\sum_{\substack{d \in \mathcal{D}  \\ R_d=r}}\min[0,\min_{\substack{l\in \mathcal{L} \\ M_{dl}=1 }}\Psi_l]
\end{align}
Observe that the minimization computed is the lowest reduced cost local assignment given a fixed global detection which is computed during each iteration of optimization.  
\end{document}